\documentclass[runningheads,a4paper]{llncs}
\usepackage{amssymb}
\usepackage{graphicx}
\usepackage{amssymb}
\usepackage[utf8]{inputenc}
\usepackage{url}
\usepackage{float}
\usepackage{amsmath}
\usepackage{graphicx}
\usepackage{wrapfig}
\usepackage{booktabs}
\usepackage{multirow}
\usepackage{url}
\usepackage{lipsum}

\begin{document}

\title{Hibikino-Musashi@Home\\2017 Team Description Paper}

\author{Sansei Hori \and Yutaro Ishida \and Yuta Kiyama \and Yuichiro Tanaka \and \\ Yuki Kuroda \and Masataka Hisano \and Yuto Imamura \and Tomotaka Himaki \and \\ Yuma Yoshimoto \and Yoshiya Aratani \and Kouhei Hashimoto \and \\ Gouki Iwamoto \and Hiroto Fujita \and Takashi Morie \and Hakaru Tamukoh }
\institute{Graduate school of life science and systems engineering,\\Kyushu Institute of Technology, \\
2-4 Hibikino, Wakamatsu-ku, Kitakyushu 808-0196, Japan, \\
\texttt{http://www.brain.kyutech.ac.jp/\~{}hma/wordpress/}}
\authorrunning{Sansei Hori et al.}
\maketitle


\begin{abstract}

Our team Hibikino-Musashi@Home was founded in 2010. It is based in Kitakyushu Science and Research Park, Japan.
Since 2010, we have participated in the RoboCup@Home Japan open competition open-platform league every year.
Currently, the Hibikino-Musashi@Home team has 24 members from seven different laboratories based in the Kyushu Institute of Technology.
Our home-service robots are used as platforms for both education and implementation of our research outcomes.
In this paper, we introduce our team and the technologies that we have implemented in our robots.

\end{abstract}


\section{Introduction}
Our team Hibikino-Musashi@Home was founded in 2010 and is based in the Kitakyushu Science and Research Park, Japan. It comprises 24 members from seven laboratories of the Kyushu Institute of Technology.
It competes in the RoboCup@Home Japan Open, an open-platform league (OPL), every year. 
We are currently developing a home-service robot and planning to use this event to present the outcomes of our research. In 2015 and 2016, we were placed third and second in the league, respectively. In addition, in 2016, we were awarded the first prize in Intelligent Home Robotics Challenge, which is a competition that takes place in Japan. This competition included a manipulation and object-recognition test and a speech-recognition and audio-detection test. These tests are similar to the RoboCup@Home competition. \par

We have three objectives.
The first is to participate in RoboCup@Home. We consider it a very important event because it gives us a chance to exhibit our research outcomes to other robot developers and research communities.
The second objective is to develop a platform for research implementation.The members of Hibikino-Musashi@Home have a variety of research backgrounds because they come from seven different laboratories. As a result, Hibikino-Musashi@Home is able to merge the research results from these laboratories and test merged systems on robots. The third objective is to develop a platform for lectures. The Kitakyushu Science and Research Park has a Joint Graduate School Intelligent Car \& Robotics Course \cite{carrobo} that is offered to both engineers and researchers human-resource development.
Robots are used as one of the educational tools in this lecture program.
Our team members can learn hardware- and software-development skills, collaboration skills that allow students to work with people outside of their own area of expertise, and team work.

This paper explains the hardware specifications and software systems as well as our scientific contributions to home-service robots.

\section{Hardware}

\subsection{Overview of our robot platforms}
We use two robots: Exi@ and Human Support Robot (HSR) \cite{toyota_hsr}.  Exi@ was developed by Hibikino-Musashi@Home in 2010. HSR was developed by the Toyota Motor Corporation, and it has been in our team since 2016. Exi@ and HSR compete in the OPL and the standard platform league (SPL), respectively.

Each robot has different characteristics that prove advantageous. Exi@, which is bigger than HSR, can perform tasks at the same scale as humans. Furthermore, all the systems that Exi@ uses have been developed by Hibikino-Musashi@Home. We have a thorough understanding of this robot, which enables us to implement our research outcomes to it and apply new devices to it. Conversely, HSR is tiny and can be easily handled. Additionally, it is equipped with high-perfection hardware, and its behavior can be quite sophisticated.

In the following section, we introduce the basic hardware information for Exi@ as well as its development history.


\subsection{Exi@}
\begin{figure}[tb]
\begin{center}
\includegraphics[width=9cm]{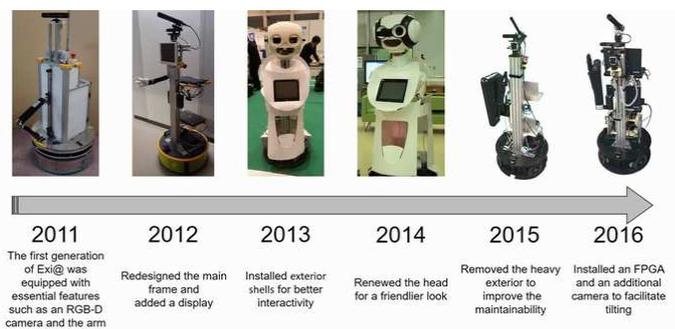}
\caption{Evolution of Exi@ from 2011 to 2016.}
\label{fig:Exi@}
\end{center}
\end{figure}

Figure \ref{fig:Exi@} shows the appearance and history of Exi@ from 2011 to 2016. The first-generation Exi@ had one RGB-D camera, an arm, and a laser range finder (LRF) mounted on a robot base. In the 2012 model, we redesigned Exi@; that design was almost the same as the present design. The 2013 and 2014 models were given exterior shells to elicit better interactivity; these models were difficult to maintain and were very heavy. Thus, it became difficult to improve Exi@ further. Consequently, the exterior had to be removed in the 2015 model. 
In the 2016 model, a vertical linear actuator camera was installed so that objects could be detected from the best position. A field-programmable gate array (FPGA) was also installed in this model: it was added to handle intelligent processing requirements.

\section{Software}
\subsection{Overview of the software}
\begin{figure}[tb]
\begin{center}
\includegraphics[width=9cm]{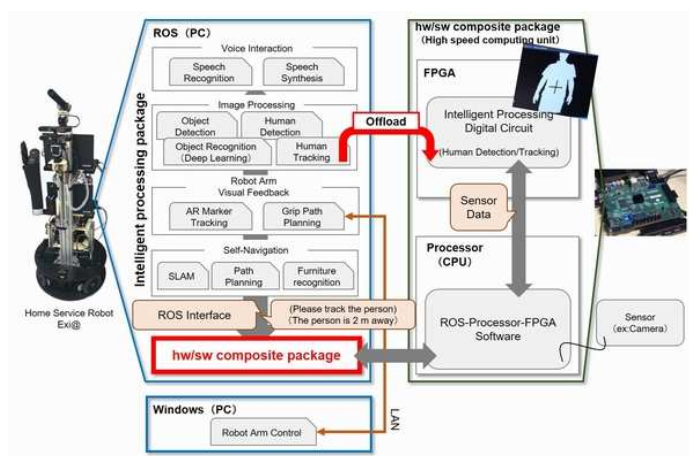}
\caption{Overview of the Exi@ software system.}
\label{fig:softOverview}
\end{center}
\end{figure}

Figure \ref{fig:softOverview} shows the software system used in Exi@. This system is based on the Robot Operating System (ROS) \cite{ros}. ROS is one of the robot middleware, and it is the de facto standard middleware for the robot systems.
Each software node used in Exi@ and HSR, e.g., image processing and manipulator control, uses an ROS interface to communicate with other software programs integrated in the robot system.
For Exi@, we have developed a number of systems, including image-processing, object-recognition, arm-control (for grabbing objects), simultaneous localization and mapping (SLAM), person-following, and sound-interaction systems, as well as an ROS-FPGA collaborative system.

\subsection{Object recognition}
The process of the object-recognition system includes:

\begin{enumerate}
\item Obtaining images of objects using the RGB-D camera and the point cloud library (PCL)\cite{pcl}, which specialize in two-dimensional (2D) / three-dimensional (3D) point cloud processing.
\item Recognizing the object in the picture from the previous step using deep learning \cite{hinton2006fast}.
\end{enumerate}

Hibikino-Musashi@Home uses a deep learning framework called Caffe \cite{caffe}. This allows us to readily develop a deep neural network. The architecture of the neural network that our system uses is a 22-layered convolutional neural network called GoogLeNet\cite{szegedy2015going}. However, the size of this network is too large for training all parameters used in a personal computer. Thus, we use transfer learning, which only trains the final layer of the network; in this training, we use images of the objects used in the RoboCup competition.

In the RoboCup@Home Japan Open, a robot must be able to recognize 15 objects. Before the competition, we create a dataset that trains the deep neural network.
We take 2,700 images of each object from various angles, and various amounts of noise are added to these images to improve our robots' object recognition accuracy.

\subsection{Manipulator control}
Exi@ is equipped with a manipulator produced by EXACT Dynamics known as iARM \cite{iarm}. This manipulator is based on the premise that it is controlled by a human user operating a wheelchair. Therefore, the positioning accuracy of the iARM is not sufficiently good.

An augmented reality (AR) marker is attached to the arm, and an AR tracker is used in our system to improve control accuracy. The AR marker tracking system is provided by ar\_track\_alvar \cite{arMarker}, which is one of the ROS packages. Arm control is performed as follows:

\begin{enumerate}
\item Obtain the coordinates of the object using PCL and deep learning.
\item Move the arm into the vicinity of the object.
\item Use ar\_track\_alvar to detect the marker on the end effector using the RGB-D camera; feedback control is then performed to minimize the error with respect to the target coordinates.
\end{enumerate}

\subsection{Voice interaction}
Voice interaction includes two systems that act as a voice recognition system and as a speech synthesis system. Our system uses rospeex \cite{rospeex}, which is a cloud service for the voice interaction. In addition, the Hibikino-Musashi@Home robots require both Japanese and English voice interaction systems so that it can use the system that situation demands. 

In the voice recognition phase, ambient noise blocks out a speaker's voice. To improve the voice-recognition accuracy, Exi@ uses a directional microphone and a microphone pan-tilt system that uses two servomotors; when listening to a speaker, the microphone turns toward it.

In a situation where a robot is being spoken to by someone whose location is unknown, the robot has to estimate the location of the speaker; our system uses HARK\cite{hark} and a microphone array to realize this.

\subsection{ROS-FPGA system}
Exi@ has a lot of intelligent systems implemented by software. Our robot has two lap-top computers to process the software demands required by these systems. 
However, these intelligent systems require a vast amount of computational resources and real-time processing to achieve a smooth interaction between Exi@ and  humans.
Thus, we installed an FPGA in Exi@. This allows some of the intelligent systems to be offloaded, which improves the overall processing speed. This is a big advantage for Hibikino-Musashi@Home when computing.

The advantages of FPGAs are that their internal digital circuit can be reconfigured and they process high parallelism, and consume a low amount of power. These advantages are ideal for the embedded systems used by home-service robots. However, the development period of FPGAs is significantly longer than that of software even if it is done by professional digital-circuit engineers.
Developing an interface that can work with both ROS and FPGA is particularly difficult; therefore, using FPGAs can be problematic for robotics engineers who also wish to develop software.

To overcome difficulty of this issue, we have proposed a novel interface called ``connective object for middleware to accelerator (COMTA)." In our system, a hardware/software (hw/sw) complex system comprising an embedded central processing unit (CPU) and an FPGA, provided by ZedBoard, is connected to a personal computer that executes ROS. 

COMTA provides a common interface that enables easy access to FPGAs from ROS. Using this, we have been able to implement a person detection and following system into the FPGA that is connected to the main system by COMTA. In this experiment, we aim to improve its electrical efficiency and processing speed.


\section{Scientific contributions}
Currently, we are attempting to implement our research outcomes into our robots so that we can present them in the Final of RoboCup@Home Japan Open. In the Final, five highest-scoring teams in the competition demonstrate their systems. In this chapter, we introduce some of our technologies that have been demonstrated in the Final.

\subsection{Operating a robot using brain waves}
In RoboCup@Home Japan Open 2015, we proposed and demonstrated a method through which a robot could be commanded using brain waves.
A user can command the robot using voice interactions or gestures, as is normally done; however, such actions are difficult for people who are unable to move, e.g., patients. Using brain waves to communicate with a robot would allow for a new interaction method.

\begin{figure}[tb]
\begin{center}
\includegraphics[width=6.5cm]{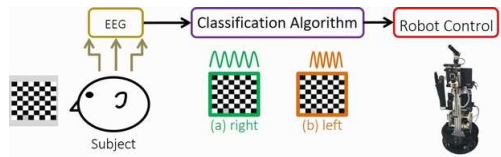}
\caption{System overview of the robot controller operated using brain waves.}
\label{fig: brainwave}
\end{center}
\end{figure}

Figure \ref{fig: brainwave} shows an overview of the system we used for this. In our system, steady-state visual evoke potential (SSVEP) was used to control Exi@. The SSVEP wave frequency was changed by a subject observing the blink pattern displayed on the monitor, as shown in Fig. \ref{fig: brainwave} (a) and (b).  We showed the subject two blink patterns that had different frequencies and then measured the SSVEP frequencies using an electroencephalograph (EEG). The robot was able to switch between the two actions which move to the right or left using the SSVEP frequency.
In the future, this could be used as a system that patients can use to call for help.

\subsection{Abnormality detection using a non-contact biosensor system}
\begin{figure}[tb]
\begin{center}
\includegraphics[width=6.5cm]{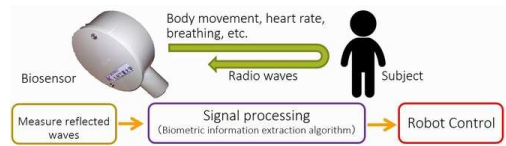}
\caption{Overview of the abnormality detection system.}
\label{fig: biosensor}
\end{center}
\end{figure}

In RoboCup@Home Japan Open 2016, we demonstrated a system that could monitor a person's health; this system used a non-contact biosensor that was an outcome of research that had been conducted at our university. Figure \ref{fig: biosensor} shows an overview of the system. In our demonstration, a robot measured the motion of a person's body and detected if they were falling. When the robot determines that a person is falling, it asks if the person is all right.

The non-contact biosensor that we installed irradiates the subject with radio waves and measures the reflection of the waves to measure vital information such as body motion, heart rate, and respiration rate. In addition, this sensor is able to measure this information through walls because it transmits radio waves. Therefore, the robot is able to monitor a subject located at a different place, e.g. if the robot is outside a bath-room that a user is currently occupying.

\vspace{-0.3cm}
\section{Conclusions}
In this paper, we introduced the basic information and objectives of Hibikino-Musashi@Home as well as the technologies used by the team. 
We have not only developed applications for the RoboCup@Home competition but have also developed applications for robots that are based on the research outcomes of our institute.
We intend to continue innovating deep learning technologies and the ROS-FPGA interface, that is COMTA.

\vspace{-0.3cm}
\section*{GitHub}
Source codes of our systems are published on GitHub. The URL is as follows:\\
https://github.com/hibikino-musashi-athome

\vspace{-0.3cm}
\section*{Acknowledgements}
This work was supported by Ministry of Education, Culture, Sports, Science and Technology, the Joint Graduate School Intelligent Car \& Robotics course (2012-2016), Kitakyushu Foundation for the Advancement of Industry Science and Technology (2013-2015), Kyushu Institute of Technology 100th anniversary commemoration projects, i.e., the student project (2015) and YASKAWA electric corporation project (2016),  and JSPS KAKENHI grant number 26330279.

\vspace{-0.4cm}



\newpage
\section*{Robot Exi@ Hardware Description}
In this section briefly describe the hardware of the robot

%

\begin{itemize}
	\item Name: Exi@.
	\item Base: RoboPlus EXIA
	\item Manipulators: Exact Dynamics iARM.
	\item LRF: Hokuyou UTM-30LX laser range finder.
	\item Microphone: SANKEN CS-3e Shotgun microphone.
	\item Batteries: Lead-acid battery 12V and 24V.
	\item Computer: ThinkPad PC Core-i5 4850U processor and 12GB RAM $\times$ 2.
	\item FPGA board: Xilinx ZedBoard\cite{zedboard} (FPGA + ARM processor).
	\item Height: About 1500 mm.
	\item Weight: About 80 kg.
	\item Base size: About 600 mm $\times$ 600 mm.
\end{itemize}


\section*{Robot's Software Description}
For our robot we are using the following software:


\begin{itemize}
	\item OS: Ubuntu 14.04.
	\item Middleware: ROS Indigo.
	\item State management: SMACH (ROS).
	\item Speech recognition (English):
		\begin{itemize}
			\item Intel RealSense SDK 2016 R2.
			\item rospeex.
			\item Web Speech API.
			\item IBM Watson Speech To Text.
		\end{itemize}
	\item Morphological Analysis Dependency Structure Analysis (English): SyntaxNet.
	\item Speech recognition (Japanese): Julius.
	\item Morphological Analysis (Japanese): MeCab.
	\item Dependency structure analysis (Japanese): CaboCha.
	\item Speech synthesis: Open JTalk.
	\item Sound location: HARK.
	\item Object detection: Point cloud library (PCL) or You only look once (YOLO)\cite{redmon2016you}.
	\item Object recognition: Caffe with GoogLeNet or YOLO.
	\item Human detection / tracking:
		\begin{itemize}
			\item Depth image + particle filter.
			\item OpenPose\cite{cao2017realtime}.
		\end{itemize}
	\item Face detection: SkyBiometory.
	\item AR mark tracking: ar\_track\_alivar (ROS).
	\item SLAM: slam\_gmapping (ROS).
	\item Path planning: move\_base (ROS).
\end{itemize}

\end{document}